\documentclass{article}

 \usepackage[position,preprint]{neurips_2026}

\usepackage[utf8]{inputenc} % allow utf-8 input
\usepackage[T1]{fontenc}    % use 8-bit T1 fonts
\usepackage{hyperref}       % hyperlinks
\usepackage{url}            % simple URL typesetting
\usepackage{booktabs}       % professional-quality tables
\usepackage{amsfonts}       % blackboard math symbols
\usepackage{nicefrac}       % compact symbols for 1/2, etc.
\usepackage{microtype}      % microtypography
\usepackage{xcolor}         % colors
\usepackage{tikz}
\usepackage{xcolor}
\usepackage{amsmath}
\usepackage{amssymb}
\usepackage{float}
\usepackage{tabularx}
\usepackage{array}
\usepackage[table]{xcolor}

\title{Time Series Forecasting Benchmarks Need Scenario-Grounded Stress Testing}

\author{
Yuyang Zhao$^{1}$, Lian Xu$^{2}$, and Hao Xue$^{1*}$\\
$^{1}$The Hong Kong University of Science and Technology (Guangzhou), Guangzhou, China\\
$^{2}$The University of Western Australia, Perth, Australia
\\
{\small $^{*}$Corresponding author: haoxue@hkust-gz.edu.cn}
}

\begin{document}

\maketitle

\begin{abstract}
% Time series forecasting (TSF) increasingly drives decisions in energy, finance, healthcare, and infrastructure, yet current evaluation remains overly narrow: benchmarks reward low held-out error, while robustness studies reduce failure to Gaussian noise, random masking, or bounded adversarial perturbations. Although real-system failures often originate from input-side anomalies, such failures cannot be characterized merely as higher average input error or noisier inputs. Such failures arise when events reshape temporal patterns, disrupt cross-variable dependencies, alter operating regimes, or propagate from faulty sensors to downstream decisions. These structured failures cannot be faithfully reproduced by existing i.i.d. perturbations, and unauditable foundation-model pretraining corpora make held-out generalization alone increasingly unreliable. We therefore advocate scenario-grounded stress testing: each test instance should include historical inputs and future targets, together with a semantic scenario, an explicit failure operator, and a measurable difficulty level. This shift makes evaluation interpretable, attributable, and deployment-relevant, enabling the community to ask not only which model is accurate, but under what conditions it fails and why it fails.

Time series forecasting (TSF) increasingly drives decisions in transportation, energy, finance, healthcare, and infrastructure, yet current evaluation remains overly narrow: standard benchmarks reward low held-out error, while robustness studies typically reduce failure to Gaussian noise, random masking, or bounded adversarial perturbations. This obscures the real failure modes of deployed forecasting systems. Input-side anomalies are not merely noisier inputs: they often reflect structured events that alter temporal dynamics, break cross-variable dependencies, induce regime shifts, or propagate from faulty sensors to downstream decisions. These semantic, causal, and system-level failures cannot be faithfully captured by i.i.d. perturbations alone. The rise of TSF foundation models makes this evaluation gap more urgent, as unauditable pretraining corpora make held-out generalization increasingly unreliable. We therefore advocate \textbf{scenario-grounded stress testing}. Each test instance should include historical inputs and future targets, together with a semantic scenario, an explicit failure operator, and a measurable difficulty level. This shift makes evaluation interpretable, attributable, and deployment-relevant and friendly, enabling the community to ask not only which model is accurate, but under what conditions it fails and why.
\end{abstract}

\section{Introduction}
Time series forecasting (TSF) lies at the operational core of energy dispatch~\citep{hong2016probabilistic,nowotarski2018recent}, financial risk management~\citep{sezer2020financial}, clinical monitoring~\citep{morid2023time,shickel2017deep}, and infrastructure planning~\citep{vlahogianni2014short}. In these domains, a forecast is not an end product but an input to a consequential downstream action. The reliability of those actions rests on an assumption that current benchmarks rarely test: errors observed during evaluation reflect the errors a model will produce after deployment.

Mounting evidence shows this assumption fails systematically rather than incidentally. When Winter Storm Uri struck Texas in February 2021, load forecasting models with low error on held-out data faced a compound disruption,record heating demand, generation freeze-offs~\citep{busby2021cascading,ferc2021february}, and correlated sensor outages in causal sequence, causing a days-long blackout affecting millions of people. The COVID-19 lockdowns offered an analogous lesson: well-calibrated demand models produced errors several times their typical magnitude as consumption patterns underwent sustained structural breaks~\citep{ruan2020cross,obst2021adaptive,farrokhabadi2022day}. 
These cases are often treated as tail events. We argue that their importance is not only their rarity, but their structure. They reshape temporal patterns, disturb dependencies between variables, change operating regimes, and allow errors in sensing to propagate into downstream decisions. Current robustness tests, which mainly perturb clean data with simple distributional noise, do not represent these mechanisms.

% Such episodes are routinely dismissed as tail events. Yet they share a distinctive structure—events that reshape temporal patterns, disrupt cross-variable dependencies, alter operating regimes, or propagate sensor faults downstream—that the current paradigm, built on i.i.d. clean-distribution perturbations, cannot represent.

\begin{figure}[H]
\centering
\includegraphics[width=\linewidth]{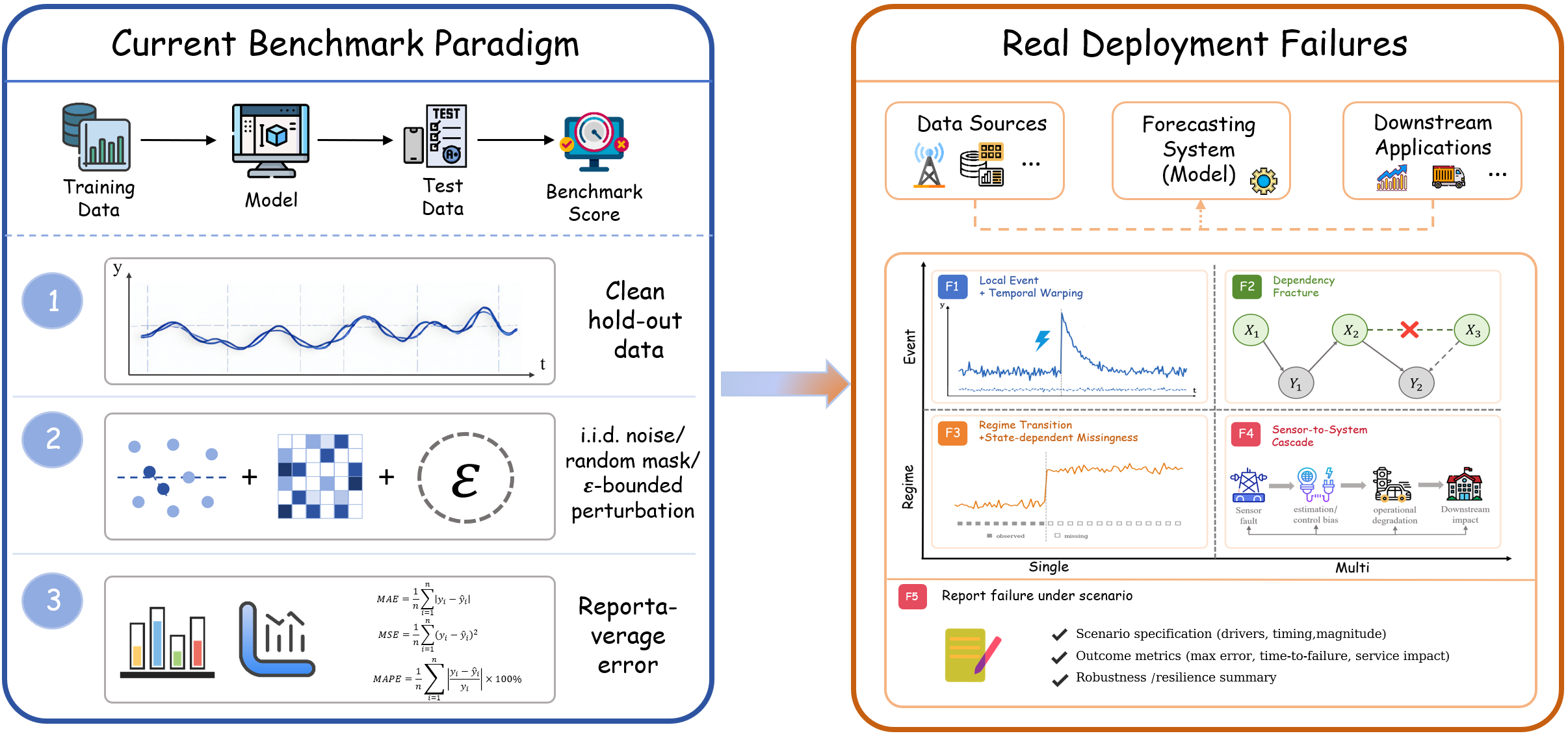}
\caption{The misalignment between current TSF benchmarking and real deployment failure. The left side evaluates models on clean held-out data with i.i.d.\ noise, masking, or $\epsilon$-bounded perturbations and reports aggregate error. The right side fails through structured events (e.g., local shocks, dependency fractures, regime transitions, cascading failures)demanding scenario-aware reporting. }
\label{fig:teaser}
\end{figure}

Mainstream TSF evaluation is still organized around a narrow question: which model achieves the lowest average prediction error on held-out data? Robustness studies extend this question by injecting Gaussian noise~\citep{cheng2024robusttsf}, random masking~\citep{du2023saits}, or bounded adversarial perturbations~\citep{liu2022practical}. While useful for testing sensitivity to local input perturbations, these settings rely on a limiting premise: failure is modeled as a value-level deviation from otherwise nominal observations, without reference to the event, system state, or causal process that generated it. In particular, i.i.d. perturbations do not encode temporal evolution, cross-variable dependency changes, or mechanisms through which errors propagate across a forecasting pipeline. A model can therefore appear robust under standard perturbation tests while remaining unevaluated under deployment-relevant failure modes. As illustrated in Figure~\ref{fig:teaser}, the gap is not merely a matter of model scale, data volume, or benchmark difficulty, but a misalignment between what benchmarks measure and what deployment demands.
% tools sharing a common premise that deviations from nominal conditions are independent, identically distributed, and devoid of causal structure. The premise is analytically convenient but, by construction, incapable of generating the failure modes described above: a perturbation drawn i.i.d.\ from a noise distribution admits no temporal shape, no cross-variable causal ordering, and no mechanism by which an error propagates through a system. The gap between benchmark performance and deployment behavior, illustrated schematically in Figure~\ref{fig:teaser}, is therefore not a matter of degree to be closed by larger models or cleaner data, but a misalignment between what benchmarks measure and what deployment demands.

\textbf{Our Position.} \textit{We argue in this position paper that the TSF community must shift its benchmarking paradigm from distributional noise to scenario-grounded stress testing.} The central claim is not that another benchmark dataset is needed, but that the object of evaluation itself must change: A next-generation TSF benchmark should characterize model behavior under parameterized, causally interpretable failure scenarios, with the parameters describing each failure reported alongside prediction error. Each test instance, under this view, is no longer an input--output pair but a structured object that reveals the conditions under which a model fails, not only the magnitude of failure.

\textbf{Contributions.} This paper makes four contributions. (I) We identify five structural blind spots in current TSF benchmarks: the absence of semantically shaped local events, of cross-variable dependency disturbances, of coupled regime-change and missingness, of cascading propagation, and of exposed scenario parameters, and argue that they are not disjoint oversights but joint projections of properties shared by real failures. 
% \textit{Second}, 
(II) We formalize scenario-grounded stress testing as a framework in which each test instance is specified by an explicit failure operator with semantically meaningful, compositional, causal, and controllable parameters. 
% \textit{Third}, 
(III) We organize meaningful TSF failures into four structural families along two orthogonal axes, i.e., event versus regime, and single-variable versus multivariate. 
% \textit{Fourth}, 
(IV) We articulate six design principles distinguishing scenario-grounded benchmarks from accuracy-centric ones, and engage three substantive counterarguments.

\section{Why Current TSF Benchmarks Fall Short}
Before arguing for an alternative paradigm, we examine why the current one produces the misalignment described above. We trace the issue to a single structural commitment shared across fifteen years of benchmark design. We then survey existing robustness research, identify the uncovered region of failure space, and derive the five structural blind spots that follow.

\subsection{Definition of the Current Paradigm}
Current TSF evaluation can be summarized as:

\begin{equation}
\mathrm{eval}(\hat{f}) \;=\; \mathbb{E}_{(X,Y)\sim \mathcal{D}_{\text{test}}}
\!\left[\ell\!\left(\hat{f}(X),\, Y\right)\right],
\label{eq:eval}
\end{equation}

where $\mathcal{D}_{\text{test}}$ is assumed either to coincide with the training distribution or to differ from it by mild covariate shift, $\hat{f}$ denotes the forecasting model, and $\ell$ is a loss function. Research progress is measured by the average reduction of this expectation across benchmark datasets.

\subsection{The Evolution of TSF Benchmarking}
Over the past fifteen years, TSF benchmarks have evolved along a coherent trajectory: 
each generation has patched limitations of its predecessor while optimizing the same 
underlying object. Table~\ref{tab:generations} summarizes this progression.

\begin{table}[H]
\centering
\caption{Generational evolution of TSF benchmarks. G1--G4 share a common logic of 
optimizing average performance on a chosen distribution; G5 (this work) reorients 
evaluation around parameterized failure scenarios.}
\label{tab:generations}
\small
\begin{tabular}{llp{5.8cm}}
\toprule
\textbf{Generation} & \textbf{Representative Benchmarks}
& \textbf{Treatment of Distribution Shift} \\
\midrule
G1 (2010s)       & M3, M4
& Implicit, via domain diversity \\
G2 (2018--2022)  & Monash, LTSF
& Explicit, as a domain-shift experiment \\
G3 (2023--2024)  & GIFT-Eval, TFB, BasicTS+
& Controlled by held-out discipline \\
G4 (2024--2025)  & fev-bench, TSFM-Bench, ProbTS
& Acknowledged but unmitigated \\
\midrule
\textbf{G5 (this work)} & \textbf{Scenario-grounded}
& \textbf{Treated as the primary design variable} \\
\bottomrule
\end{tabular}
\end{table}

The first generation (M3, M4) addressed forecast accuracy across heterogeneous domains~\citep{makridakis2000m3,makridakis2020m4}, encoding distributional variability implicitly through domain diversity. The second generation (Monash~\citep{godahewa2021monash}, LTSF~\citep{zhou2021informer,wu2021autoformer,nie2022time}) made domain shift an explicit experimental condition. 
The third generation (GIFT-Eval~\citep{aksu2024gift}, TFB~\citep{qiu2024tfb}, BasicTS+~\citep{shao2024exploring}) responded to the rise of pretrained forecasters by attempting to control distributional contamination through community discipline around held-out datasets. The fourth generation (fev-bench~\citep{shchur2025fev}, TSFM-Bench~\citep{li2025tsfm}, ProbTS~\citep{zhang2024probts}) acknowledged that pretraining leakage can no longer be reliably excluded in the foundation-model era and shifted toward more realistic evaluation protocols. Yet across 
all four generations, the object being measured has remained Equation~\eqref{eq:eval}: 
average error on some distribution. We argue that the next generation must change this 
object itself, treating structured failure conditions not as a confound to be isolated 
but as the principal subject of evaluation.

\subsection{Existing Robustness Research}

TSF robustness has been extensively studied, yet existing literature occupies three of the four quadrants defined by two axes: deviations are either sampled (from a distribution) or constructed (by an explicit operator), and either unstructured (i.i.d. across time and variables) or structured (temporal shape, cross-variable coupling, causal propagation). Table~\ref{tab:quadrants} summarizes prior work across quadrants.

\begin{table}[H]
\centering
\caption{Four quadrants of TSF robustness research, organized by how deviations are 
generated (sampled vs.\ constructed) and whether they carry temporal, cross-variable, 
or causal structure.}
\label{tab:quadrants}
\small
\renewcommand{\arraystretch}{1.6}
\begin{tabular}{>{\bfseries}r p{5.5cm} p{5.5cm}}
\multicolumn{1}{c}{} & \multicolumn{1}{c}{\textbf{Sampled}} & \multicolumn{1}{c}{\textbf{Constructed}} \\ \cline{2-3}
Unstructured & 
\cellcolor{blue!5} Noise \& Missingness \newline \scriptsize (Gaussian, uniform, random masking) & 
\cellcolor{red!5} Adversarial Attacks \newline \scriptsize (FGSM, PGD, C\&W within $\epsilon$-balls) \\ 
Structured & 
\cellcolor{orange!5} Distribution Shift \newline \scriptsize (Wild-Time, temporal drift) & 
\cellcolor{green!10} \textbf{Scenario-Grounded} \newline \textbf{\scriptsize (this work)} \\ 
\end{tabular}
\end{table}

Existing robustness work has shown that forecasting models can be brittle under small worst-case 
perturbations, Gaussian or uniform noise, random masking, and temporal distribution shift. 
Yet these lines of work remain limited in distinct but related ways: noise and 
masking~\citep{cheng2024robusttsf,du2023saits,tashiro2021csdi} assume i.i.d.\ timesteps 
and cannot express cross-temporal failure structure; adversarial 
perturbations~\citep{liu2022practical} are mathematical objects without physical meaning 
and lack semantic grounding; and distribution-shift 
evaluation~\citep{yao2022wild,gagnon2022woods,kim2021reversible}, though closest to our 
concern, remains observational, measuring degradation across time gaps without enabling 
counterfactual control over how the same shift would behave if made more severe. As a 
result, these approaches can reveal fragility, but they cannot provide the experimentally 
controlled, causally interpretable attribution required for deployment-grade reliability 
assessment. 

The uncovered quadrant (bottom-right) contains failures that are both structured and constructed, precisely the failure mode described in Section 1 and 
detailed below in Section 2.4. These dominate real deployments yet have 
no established evaluation methodology. Foundation-model 
benchmarking~\citep{aksu2024gift,shchur2025fev,liang2024foundation} is a complementary 
concern that asks whether reported numbers can be trusted; the present work argues that 
filling this quadrant is what makes such trust meaningful in the first place.

\subsection{Structural Blind Spots in Real TSF Failures}
Documented TSF failures across domains, including the ERCOT breakdown during Winter Storm Uri~\citep{busby2021cascading,ferc2021february}, the multi-fold degradation of electricity demand models during COVID-19~\citep{ruan2020cross,obst2021adaptive,farrokhabadi2022day}, the documented brittleness of clinical event-prediction models in ICU sepsis monitoring~\citep{moor2021early,seymour2017time}, and the cascading line trips of the 2006 European blackout~\citep{li2007analysis}, share five recurring properties: temporal shape, cross-variable structure, mechanism coupling, causal propagation, and attributability. These are not an arbitrary feature list but ontological properties that distinguish structured failure from unstructured noise; removing any one yields a case still statistically sampleable but no longer a real deployment failure. Current TSF benchmarks lack mechanisms to represent these properties. Table~\ref{tab:blindspots} maps each property to its corresponding structural blind spot; we elaborate on each in turn.

\begin{table}[H]
\centering
\caption{Five ontological properties of real TSF failures and the corresponding structural blind spots in the current evaluation paradigm.}
\label{tab:blindspots}
\small
\renewcommand{\arraystretch}{1.6} % 进一步增加行高，配合宽版布局
\rowcolors{2}{gray!10}{white}
\begin{tabularx}{\textwidth}{c l X}
\toprule
\rowcolor{gray!25} 
\textbf{ID} & \textbf{Ontological Property} & \textbf{Blind Spot in Current Evaluation} \\
\midrule
\textbf{B1} & Temporal shape           & No semantically shaped local events \\
\textbf{B2} & Cross-variable structure  & No cross-variable dependency disturbances \\
\textbf{B3} & Mechanism coupling       & No coupling of regime change and missingness \\
\textbf{B4} & Causal propagation       & No cascading failure propagation \\
\textbf{B5} & Attributability          & No exposure of scenario parameters \\
\bottomrule
\end{tabularx}
\end{table}

\textbf{B1: No semantically shaped local events.} PGD perturbations~\citep{madry2017towards} are scattered points within an $\epsilon$-ball; i.i.d.\ noise forbids temporal correlation by definition. Neither admits the notion of temporal shape, onset, peak, or decay cannot exist within a probability model over independent draws, despite the long-recognized importance of event-shaped anomalies in time series~\citep{keogh2005hot,chandola2009anomaly}.

\textbf{B2: No cross-variable dependency disturbances.} Evaluation on multivariate benchmarks~\citep{lai2018modeling,wu2022timesnet} is routinely reduced to independent univariate assessments~\citep{nie2022time} or noise is applied isotropically across dimensions. The pair format $(X, Y)$ thus has no addressable position for failures in which each series appears normal while the inter-variable structure is pathological, a configuration well documented in studies of cross-asset financial contagion~\citep{forbes2002no}.

\textbf{B3: No coupling of regime change and missingness.} Change-point detection~\citep{aminikhanghahi2017survey} and missing-data imputation~\citep{cao2018brits,du2023saits} have developed in isolation, despite the well-documented joint occurrence of these phenomena in stressed systems~\citep{little2019statistical}. Treating them as independent axes misrepresents their joint distribution.

\textbf{B4: No cascading failure propagation.} The standard format assumes synchronous samples, faults as input noise, and targets from a normally operating system, making dependency-chain error propagation inexpressible, since what propagates is the fault itself, not i.i.d. input corruption. Cascading failures are extensively studied in power-grid reliability~\citep{dobson2007complex,buldyrev2010catastrophic} and supply-chain disruption, yet remain absent from TSF benchmarks.

\textbf{B5: No exposure of scenario parameters.} Even with hard samples, the reason for difficulty remains implicit: a reported MAE cannot be decomposed into shock magnitude, temporal distortion, or cross-variable coupling. B5 is the meta-level blind spot, even generative solutions to B1--B4 would not enable attribution without exposing scenario parameters. The need for transparent reporting of evaluation conditions has been emphasized in model documentation frameworks~\citep{mitchell2019model,gebru2021datasheets}, yet remains rarely implemented in TSF benchmarks.

% \textbf{Our Position.} 
The misalignment between benchmark and deployment in TSF is not a measurement gap but a representational one. The five blind spots B1--B5 are not features missing from current benchmarks; they are properties of real failure that the current paradigm's representational scheme, i.i.d.\ perturbations of a clean distribution, is structurally unable to express. Filling the uncovered quadrant, therefore, requires not a richer noise model but a different object of evaluation, one in which the structure of the failure condition is part of what each test instance is, not merely a property of how it was sampled. The next section advances three arguments for why this reconception is necessary rather than merely desirable.

\section{Three Arguments for a Paradigm Shift}
We now advance three cumulative arguments for why the representational commitment identified in Section 2.4 must be replaced rather than patched from within.

\subsection{Argument I: Structured Failures Are Not Expressible as i.i.d.\ Perturbations}
The obstacle is not that i.i.d.\ perturbations are too simple but that the five ontological properties underlying B1–B5 impose requirements on the generative process that independence cannot satisfy. Temporal shape requires dependence across timesteps; cross-variable structure requires directed coupling across dimensions~\citep{mcdonald2002judea}; mechanism coupling requires shared causes across failure channels; causal propagation requires a fault to move through a dependency chain; and attributability requires that the generating parameters themselves be exposed as part of the test instance. What unifies these is that each concerns how a test instance is constructed, not merely which numerical values it contains, an instance of the broader principle that interventional and counterfactual reasoning are irreducible to observational sampling~\citep{mcdonald2002judea,scholkopf2021toward}. Making B1--B5 expressible, therefore, requires a minimal change in the generative contract: from sampling-from-a-distribution to constructing-via-an-operator, with the operator and its parameters $\Theta$ promoted to first-class objects reported alongside each instance.

\subsection{Argument II: Held-Out Evaluation Has Lost Its Epistemic Guarantee}
Held-out evaluation rests on a simple epistemic guarantee: unseen test data makes low test error interpretable as generalization. In the foundation-model era, this guarantee weakens as TSFMs are pretrained on large, partially unauditable corpora~\citep{ansari2024chronos,das2023decoder,woo2024unified,liu2025sundial,goswami2024moment} drawn from the same public repositories that later supply benchmark test sets. As the boundary between pretraining and testing becomes uncertain, strong performance becomes difficult to distinguish from recall.

Embargoes, time cutoffs, and private test sets are necessary but reactive measures, since released benchmarks can enter future pretraining cycles. Scenario-grounded evaluation offers a stronger defense: stressed instances are constructed at evaluation time by $\phi(X; \Theta)$, so even if $X$ was seen, the structured failure induced by a controlled, previously unexposed $\Theta$ was not, shifting evaluation from reproducing historical distributions to testing reliability under interpretable failure conditions.

\subsection{Argument III: Deployment Asks a Different Question Than Average Accuracy}
Even if i.i.d.\ perturbations could express structured failure and held-out data remained guaranteed unseen, the current paradigm would answer the wrong question. Deployment asks not only about average error, but under what conditions a model fails, how it fails, whether the failure propagates, and what consequences follow. These questions are governed by the failure distribution's tail, not its center.

High-stakes AI practice increasingly requires scenario-conditioned documentation of when systems fail, how severely, and with what probability~\citep{smuha2025regulation,working2025good,nerc2011reliability}. Metrics based on $\epsilon$-bounded perturbations, Gaussian noise variance, or random masking rates cannot express risk in the language used by operators, regulators, and domain experts; scenario parameters $\Theta$ can directly encode heat-wave demand surges, sustained sensor drift, or regime shifts coupled with rising missingness, making evaluation actionable rather than merely reportable.

% \paragraph{Our Position.}
Scenario-grounded stress testing is necessary because the current paradigm fails in three independent ways: it cannot represent structured failure (an ontological failure), it can no longer guarantee what its measurements mean (an epistemic failure), and it answers a question that deployment does not pose (a normative failure). Any one of these would suffice to motivate methodological reform; their conjunction makes reform inescapable.

\section{Scenario-Grounded Stress Testing: A Framework}
The three arguments in Section 3 establish what the current paradigm cannot do. This section specifies what a principled alternative requires. We formalize scenario-grounded stress testing, define the requirements that scenario parameters must satisfy to be scientifically useful, and organize the space of meaningful TSF failures into a structured taxonomy.

\begin{figure}[H]
\centering
\includegraphics[width=\linewidth]{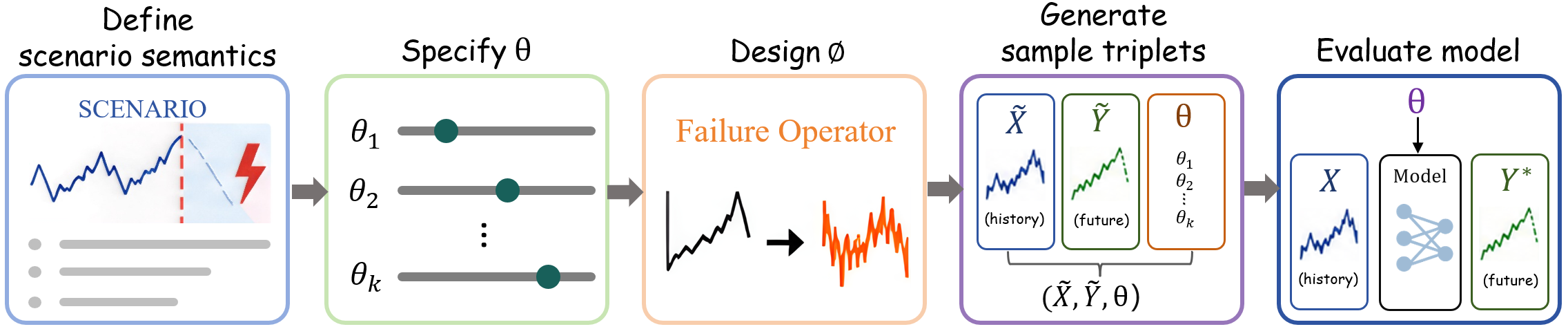}
\caption{Workflow of scenario-grounded stress testing. This workflow (i) defines the
semantic content of a target scenario, (ii) specifies parameter vector $\Theta$,
(iii) designs failure operator $\phi$, (iv) generates evaluation instances
$(\tilde{X}, \tilde{Y}, \Theta)$ by applying $\phi$ to clean base pairs $(X, Y)$, and
(v) evaluates the model with $\Theta$ exposed as conditioning information rather than
discarded as latent metadata.}
\label{fig:workflow}
\end{figure}

\subsection{Formal Definition}
The current evaluation paradigm treats a test instance as a pair $(X, Y)$, where $X \in \mathbb{R}^{T \times d}$ is the context and $Y \in \mathbb{R}^{H \times d}$ is the forecast target. We replace this pair with a structured quintet:

\begin{equation}
\text{instance} \;=\; \bigl(\,\tilde{X},\; \tilde{Y},\; \Theta,\; \phi,\; \delta\,\bigr),
\label{eq:instance}
\end{equation}
where $\phi(\,\cdot\,;\Theta)$ is an explicit \emph{failure operator} parameterized by scenario parameter vector $\Theta$, applied to a clean base pair $(X, Y)$ to produce stressed input $\tilde{X} = \phi(X;\Theta)$ and target $\tilde{Y}$ ($\tilde{Y} = Y$ when only the input window is perturbed), and the difficulty score is defined as $\delta = \kappa(\Theta)$, where $\kappa$ is a monotone, domain-interpretable mapping from scenario parameters to a scalar difficulty level. The crucial departure from Equation~\eqref{eq:eval} is that $\Theta$, $\phi$, and $\delta$ are first-class experimental quantities reported with each instance, not latent properties of a test distribution. Evaluation becomes
\begin{equation}
\mathrm{eval}(\hat{f},\,\Theta) \;=\; \mathbb{E}\!\left[\,\ell\!\left(\hat{f}(\tilde{X}),\,
\tilde{Y}\right)\,\big|\,\Theta\,\right],
\label{eq:eval-new}
\end{equation}

where conditioning on $\Theta$ makes explicit what was implicit: performance is reportable as a function of $\Theta$ rather than as an aggregate against a monolithic distribution, enabling the attribution the current paradigm precludes. Figure~\ref{fig:workflow} illustrates the workflow.

\subsection{Four Requirements on Scenario Parameters}
Not every parameterized perturbation qualifies as a scenario parameter in the sense of
Equation~\eqref{eq:instance}. We require that $\Theta$ satisfy four properties.

\textbf{Semantic interpretability.} Each component of $\Theta$ must correspond to a
quantity that a domain expert can recognize and reason about: shock magnitude, event
duration, coupling gain, sensor failure rate. A parameter interpretable only as a
coordinate in a mathematical space, such as the step size of a gradient-based
perturbation, does not qualify. This requirement ensures that evaluation results are
legible to the practitioners who must act on them, not only to the researchers who
produced them.

\textbf{Compositionality.} Scenario parameters must support structured combination.
Components of $\Theta$ should compose to express compound failure
conditions, simultaneous demand surge and sensor dropout, or a regime shift followed by
cascading propagation, without requiring a fresh parameter vector for each combination.
This property enables systematic coverage of the joint failure space rather than mere enumeration of isolated point scenarios.

\textbf{Causal grounding.} $\Theta$ must encode a mechanism rather than a statistical summary~\citep{mcdonald2002judea,peters2017elements}. A parameter that specifies the probability of a level shift is a statistical description; a parameter that specifies the trigger condition, propagation path, and decay dynamics of that shift is a causal one. Causal grounding allows a benchmark to express not only that a test instance is difficult but why, and to do so in terms a practitioner can map back to a physical, biological, or economic process.

\textbf{Difficulty controllability.} The difficulty level $\delta$ must vary monotonically
and predictably with $\Theta$, so that an experimenter can construct test batteries at
graded severity levels and attribute performance differences to a specific parameter
changes. Without this property, a benchmark may generate hard samples but cannot support
the systematic sensitivity analysis that robustness certification requires.

\subsection{A Taxonomy of Structural Failure Families}

Drawing on documented failure modes in energy systems~\citep{busby2021cascading,andersson2005causes}, financial markets~\citep{https://doi.org/10.1111/jofi.12498}, and clinical monitoring~\citep{moor2021early}, we organize the space of meaningful TSF failures along two orthogonal axes:
whether the failure manifests as a \emph{local event}, bounded in time, with onset and
offset, or a \emph{regime change}, a persistent shift in the data-generating mechanism;
and whether it involves single or multiple interacting variables.
This yields four structural families, as shown in Figure~\ref{fig:teaser} (right).

The four families are not merely a convenient organizing scheme; they constitute a
constructive minimum. Omitting F1 leaves a benchmark unable to evaluate response to
bounded perturbation events, the most frequent single-series anomaly class in practice.
Omitting F2 makes invisible the failure mode in which each series appears normal yet the inter-variable structure is pathological. Omitting F3 conflates transient
and persistent failures, precluding distinction between models that recover after a shock and those that never recalibrate after a regime shift. Omitting F4 leaves
cascading propagation, the most consequential failure mode in critical infrastructure, entirely unrepresented. A benchmark that covers all four families does
not exhaust possible failures, but it guarantees that no major structural
class is invisible. Each family admits a natural parameterization in $\Theta$, amplitudes and durations for F1, coupling structures for F2, persistence and missingness for F3, and propagation graphs for F4, providing a constructive scaffolding for instantiating $\Theta$ in practice.

\begin{table}[H]
\centering
\caption{Six design principles for next-generation TSF benchmarks. The principles operationalize the formal framework into a contract that any fifth-generation benchmark should satisfy.}
\label{tab:principles}
\small
\begin{tabular}{p{0.48\linewidth}p{0.48\linewidth}}
\toprule
\textbf{Principles P1 -- P3} & \textbf{Principles P4 -- P6} \\
\midrule

\textbf{P1. Expose scenario parameters}
Every test instance must be released with its scenario parameters $\Theta$ and
difficulty level $\delta$, so that performance can be analyzed against specific failure
conditions rather than only as an aggregate. 
\medskip

\textbf{P2. Cover structural failure families}
A benchmark must include instances from each of F1--F4, avoiding the implicit omission
of failure classes such as dependency fracture or cascading propagation.
\medskip

\textbf{P3. Vary difficulty systematically}
Each family must be instantiated across an ordered range of difficulty levels, supporting
robustness curves and threshold analyses rather than single-point comparisons.
&

\textbf{P4. Separate mechanism from observation}
Mechanism change and observation degradation must be parameterized independently, so their individual and joint contributions to forecast error are decomposable.

\medskip

\textbf{P5. Preserve base-sequence fidelity}
Failure operators must be applied to real or simulated base sequences $X$, ensuring that the difficulty introduced remains anchored to plausible deployment conditions.

\medskip

\textbf{P6. Report results attributably}
Results must be stratified by failure family, parameter, and difficulty level rather
than reported as a single aggregate, so that a benchmark indicates not only how a model
performs but where and why it fails.
\\
\bottomrule
\end{tabular}
\end{table}

\subsection{Design Principles for Next-Generation TSF Benchmarks}

The framework above motivates a set of design principles, summarized in Table~\ref{tab:principles}. We treat them not as a checklist but as an evaluation contract that any fifth-generation benchmark should satisfy.

A benchmark satisfying these principles is not one that adds harder samples to an existing test set, but one that changes what a test instance records, not only the input and target, but the conditions under which the model is being evaluated.

% \textbf{Our Position.} 
Scenario-grounded stress testing redefines the object that TSF evaluation is meant to test, not the dataset on which it is run. Scenario parameters, structural failure families, and design principles are not separate additions but mutually connected components of a single evaluation contract, one that follows directly from the need to make temporal shape, cross-variable coupling, mechanism disruption, causal propagation, and decision relevance explicitly observable in evaluation.

\section{Alternative Views and Counterarguments}
% Three objections deserve direct engagement. We address them in order of increasing subtlety.

% \subsection{``Scaling Law Will Solve This''}
\paragraph{Alternative View I: Scaling Law Will Solve This.}
% \textit{Objection.} 
The most direct response to our diagnosis holds that scale will absorb the failure modes we describe: as pretraining corpora scale to trillions of timestamps~\citep{kaplan2020scaling,hoffmann2022training}, foundation models will have seen enough Uri-like, COVID-like, and cascade-like events to forecast their successors. Scenario-grounded stress testing, on this view, addresses a transient gap that scale will close.

The argument conflates fitting past extremes with anticipating novel ones. The failures that matter are, by definition, novel relative to any past corpus; memorizing every historical extreme leaves the structured failures one has not yet seen still untested, constructing them remains a primary way to evaluate that behavior. Scale also conflates coverage with attribution: even granting correct behavior on every historical failure, the paradigm cannot reveal whether the model learned the mechanism or merely memorized the trace, a distinction reinforced by recent findings that benchmark performance masks reliance on spurious correlations~\citep{geirhos2020shortcut,d2022underspecification}. Scenario-grounded evaluation is not in competition with scaling; it is the diagnostic instrument that holds scaling accountable.

% \subsection{``Synthetic Scenarios Are Less Real Than Historical Events''}
% \textit{Objection.} 

\paragraph{Alternative View II: Synthetic Scenarios Are Less Real Than Historical Events.}
This objection reverses direction: real historical failures, however rare, are more ecologically valid than anything produced by $\phi(X;\Theta)$. A synthesized cascade is a researcher's caricature; benchmarks should be built from documented incidents, not from parameterized constructions.

The objection rightly identifies ecological validity as non-negotiable but locates it in the wrong place: validity inheres in structural fidelity, not in historical origin. A synthesized scenario that preserves the relevant structure is more ecologically valid than a historical record stripped of causal context and recorded under uncontrolled observation~\citep{rubin1974estimating,imbens2015causal}, a principle already established in safety-critical AI evaluation, where parameterized scenario generation is standard practice for autonomous driving~\citep{ding2023survey,fremont2019scenic}. The objection also overlooks a practical asymmetry: documented failures are scarce because they break systems, leaving few historical samples per family for robustness certification. Parameterized construction is therefore not a substitute for historical grounding but the mechanism that turns sparse observations into coverage-adequate evaluation.

% \subsection{``Exposed Parameters Invite Goodhart's Law''}
% \textit{Objection.} 

\paragraph{Alternative View III: Exposed Parameters Invite Goodhart's Law.}
A subtler objection accepts the proposed framework but warns of a self-undermining dynamic~\citep{goodhart1984problems,manheim2018categorizing}. Once $\Theta$ is exposed as a first-class quantity, models will be optimized against the specific $\Theta$ values the benchmark uses, and performance will rise without genuine improvement in robustness. Scenario-grounded benchmarks, on this view, will be gamed precisely because they are transparent.

The risk is real but it ignores structural defenses unavailable to accuracy-centric benchmarks. Because $\Theta$ is compositional, a benchmark can hold out specific parameter combinations during release, exposing the parameterization while testing on combinations no participant has seen, a defense that scales with the dimensionality of $\Theta$. More fundamentally, the alternative is not a benchmark immune to gaming but one whose gaming is invisible: the absence of exposed parameters does not prevent overfitting, only conceals it. Transparency is a feature, not a vulnerability.

% \textbf{Our Position.} 
\paragraph{Our Position.}
The three counterarguments share a common form: each accepts part of our diagnosis yet claims that an extended existing paradigm can absorb the reform without changing the evaluation object. None succeeds. Scale cannot provide attribution; historical fidelity cannot ensure coverage; transparency does not invite gaming but exposes it. Scenario-grounded stress testing is therefore not optional but the response each counterargument ultimately requires.

\section{Implications and Open Problems}
The position advanced in this paper redirects several active research lines in TSF, and several open problems become tractable that were previously ill-posed.

\subsection{Implications for Current Research Directions}
\paragraph{Foundation model evaluation.} Our framework reframes the central question of TSFM benchmarking: rather than racing against pretraining corpora to preserve held-out integrity~\citep{magar2022data,sainz2023nlp}, the community can adopt evaluation protocols whose validity does not depend on data novelty, an approach paralleling behavioral~\citep{ribeiro2020beyond} and dynamic-benchmarking~\citep{kiela2021dynabench} evaluation in NLP.

\paragraph{Time series anomaly detection.} The standard anomaly types (e.g., point, contextual, and collective~\citep{chandola2009anomaly,blazquez2021review}) provide natural primitives for the failure operators of F1--F4. Detection methods~\citep{audibert2020usad,xu2021anomaly} can therefore serve as failure operator constructors, while scenario-grounded benchmarks provide the standardized testbeds detection research currently lacks~\citep{wu2021current}.

\paragraph{Robust and uncertainty-aware forecasting.} Probabilistic and quantile-based methods~\citep{salinas2020deepar,rasul2021autoregressive,gneiting2007strictly} have traditionally been evaluated by calibration on held-out distributions. Under scenario-grounded evaluation, they can be tested on whether predicted uncertainty correctly anticipates the difficulty $\delta$ of structured failures, which is a stronger criterion for calibration in deployment.

\subsection{Open Problems}

\paragraph{Scenario libraries and difficulty calibration.} F1--F4 specify a structural minimum, not a concrete vocabulary. The community needs shared domain-specific catalogs of $\phi$ and $\Theta$ calibrated to documented failures. This is analogous to scenario catalogs in autonomous driving safety~\citep{fremont2019scenic,riedmaier2020survey}. Cross-domain $\delta$-calibration protocols are also needed to make domain-specific catalogs comparable.

\paragraph{Composition with foundation models.} Current TSFMs~\citep{ansari2024chronos,das2023decoder,woo2024unified,goswami2024moment} are trained as zero-shot forecasters. Their behavior under structured failure is largely uncharacterized. Stress-testing pretrained TSFMs along F1--F4 axes would provide the diagnostic basis for understanding which failure modes scale solves and which it does not, echoing emergent-capability debates in language models~\citep{wei2022emergent,schaeffer2023emergent}.

\paragraph{Rethinking evaluation metrics.} MSE, MAE~\citep{hyndman2006another,kim2016new}, and their probabilistic counterparts~\citep{gneiting2007strictly,matheson1976scoring} applied over the forecast horizon, treat errors as exchangeable, the metric expression of the commitment producing B1–B5. Scenario-grounded evaluation calls for metrics that report when an error occurs (peak-time deviation, time-to-recovery), where it propagates (cross-variable error correlation), and whether it crosses operational thresholds (exceedance probability, time-in-violation). Designing a metric vocabulary aligned with $\Theta$ is, alongside the framework, an open task.

\section{Conclusion}
Time series forecasting has matured into a discipline whose models inform decisions where failure costs money, time, and lives; its benchmarks have not. They continue to measure average error on clean distributions, while deployed systems are decided by structured failures rather than random noise. This misalignment is representational, not metric: it is rooted in what evaluation takes as its object. The question this paper leaves the community is therefore not whether to build another benchmark, but whether TSF evaluation will continue to ask only how accurate models are, or finally begin to ask, in the structured terms the answer requires: under what conditions models fail, and why.

\bibliography{main}
\bibliographystyle{plain}

%%%%%%%%%%%%%%%%%%%%%%%%%%%%%%%%%%%%%%%%%%%%%%%%%%%%%%%%%%%%

\end{document}